\documentclass[sigconf]{acmart}

\makeatletter
\def\mdseries@tt{m}
\makeatother

\usepackage{adjustbox}
\usepackage{amssymb}
\usepackage{bm}
\usepackage{caption}
\usepackage{color}
\usepackage{makecell}
\usepackage{microtype}
\usepackage{multirow}
\usepackage{paralist}
\usepackage{rotating}
\usepackage{subcaption}
\usepackage{tabularx}
\usepackage{xurl}
\usepackage{xcolor}
\usepackage{enumitem}
\usepackage{booktabs}
\usepackage{algpseudocode}
\usepackage{listings}
\usepackage{hyperref}
\usepackage{xcolor}
\usepackage{graphicx}
\usepackage[normalem]{ulem}
\usepackage[linesnumbered,ruled]{algorithm2e}
\usepackage[capitalize]{cleveref}
\usepackage{tikz,nicefrac,amsmath,pifont,tkz-graph,makecell,tkz-graph}
\usepackage[utf8x]{inputenc}
\usepackage{textgreek}
\usepackage[frozencache=true]{minted}

\usetikzlibrary{arrows,backgrounds,patterns,matrix,shapes,fit,calc,shadows,plotmarks}
\usetikzlibrary{arrows.meta}

\usepackage[english]{babel}
\PassOptionsToPackage{dvipsnames}{xcolor}
\PassOptionsToPackage{hyphens}{url}
\definecolor{mylinkcolor}{RGB}{0,0,140}
\definecolor{ForestGreen}{RGB}{34,139,34}
\hypersetup{colorlinks,allcolors=mylinkcolor,citecolor=mylinkcolor}

\newcommand{\given}{\;\vert\;}

\newcommand{\rfrac}[2]{{}^{#1}\!/_{#2}}

\newcommand{\phantomsubfigure}[1]{\begin{subfigure}[b]{0.1\textwidth}\phantomcaption\label{#1}\end{subfigure}}

\newcommand{\xhdr}[1]{\vspace{0.5mm}\noindent{\textbf{#1.}}\hspace{0.5mm}}

\newcommand{\nrm}[1]{\textcolor[rgb]{0.0,0.0,0.0}{#1}}
\newcommand{\bst}[1]{\textcolor[RGB]{27,158,119}{\textbf{#1}}}
\newcommand{\wst}[1]{\textcolor[RGB]{217,95,2}{\textbf{#1}}}

\lstdefinelanguage{Julia}%
  {morekeywords={abstract,break,case,catch,const,continue,do,else,elseif,%
      end,export,false,for,function,immutable,import,importall,if,in,%
      macro,module,otherwise,quote,return,switch,true,try,type,typealias,%
      using,while},%
   sensitive=true,%
   alsoother={\$},%
   morecomment=[l]\#,%
   morecomment=[n]{\#=}{=\#},%
   morestring=[s]{"}{"},%
   morestring=[m]{'}{'},%
}[keywords,comments,strings]%

\copyrightyear{2020}
\acmYear{2020}
\setcopyright{acmcopyright}\acmConference[KDD '20]{Proceedings of the 26th ACM SIGKDD Conference on Knowledge Discovery and Data Mining}{August 23--27, 2020}{Virtual Event, CA, USA}
\acmBooktitle{Proceedings of the 26th ACM SIGKDD Conference on Knowledge Discovery and Data Mining (KDD '20), August 23--27, 2020, Virtual Event, CA, USA}
\acmPrice{15.00}
\acmDOI{10.1145/3394486.3403101}
\acmISBN{978-1-4503-7998-4/20/08}

\begin{document}
\fancyhead{}

\title{Residual Correlation in Graph Neural Network Regression}

\author{Junteng Jia}
\affiliation{\institution{Cornell University}}
\email{jj585@cornell.edu}

\author{Austin R. Benson}
\affiliation{\institution{Cornell University}}
\email{arb@cs.cornell.edu}


\begin{abstract}
  A graph neural network transforms features in each vertex's neighborhood into a vector representation of the vertex.
  Afterward, each vertex's representation is used independently for predicting its label.
  This standard pipeline implicitly assumes that vertex labels are
  conditionally independent given their neighborhood features.
  However, this is a strong assumption, and we show that it is far from true on
  many real-world graph datasets.
  Focusing on \emph{regression} tasks, we find that this conditional
  independence assumption severely limits predictive power.
  This should not be that surprising, given that traditional graph-based semi-supervised learning methods such as label propagation work in the opposite
  fashion by explicitly modeling the correlation in predicted outcomes.
  
  Here, we address this problem with an interpretable and efficient framework
  that can improve any graph neural network architecture simply by exploiting
  correlation structure in the regression residuals.
  In particular, we model the joint distribution of residuals on vertices with a
  parameterized multivariate Gaussian, and estimate the parameters by maximizing
  the marginal likelihood of the observed labels.
  Our framework achieves substantially higher accuracy than competing baselines,
  and the learned parameters can be interpreted as the strength of correlation among connected vertices.
  Furthermore, we develop linear time algorithms for low-variance, unbiased model parameter estimates,
  allowing us to scale to large networks.
  We also provide a basic version of our method that makes stronger
  assumptions on correlation structure but is painless to implement,
  often leading to great practical performance with minimal overhead.
\end{abstract}

\maketitle
%
\section{Exploiting Residual Correlation \label{sec:intro}}
Graphs are standard representations for wide-ranging complex systems with interacting entities,
such as social networks, election maps, and transportation systems~\cite{Newman_2010,Easley_2010,Fernadndez_2014}.
Typically, a graph represents entities as vertices (nodes) and the
interactions as edges that connect two vertices.
An attributed graph further records attributes of interest for each vertex;
for example, an online social network may have information on a person's location, gender, and age.
However, some attribute information might be missing on a subset of vertices.
Continuing our online social network example, some users may skip gender during survey or
registration, which one may want to infer for better targeted advertising.
Or, in U.S.\ election map networks, we may have polling data from some counties and
wish to predict outcomes in other ones, given commonly available demographic information
for all the counties.

\begin{figure}[t!]
\phantomsubfigure{fig:intro_a}
\phantomsubfigure{fig:intro_b}
\phantomsubfigure{fig:intro_c}
\phantomsubfigure{fig:intro_d}
\phantomsubfigure{fig:intro_e}
\centering
\includegraphics[width=1.00\linewidth]{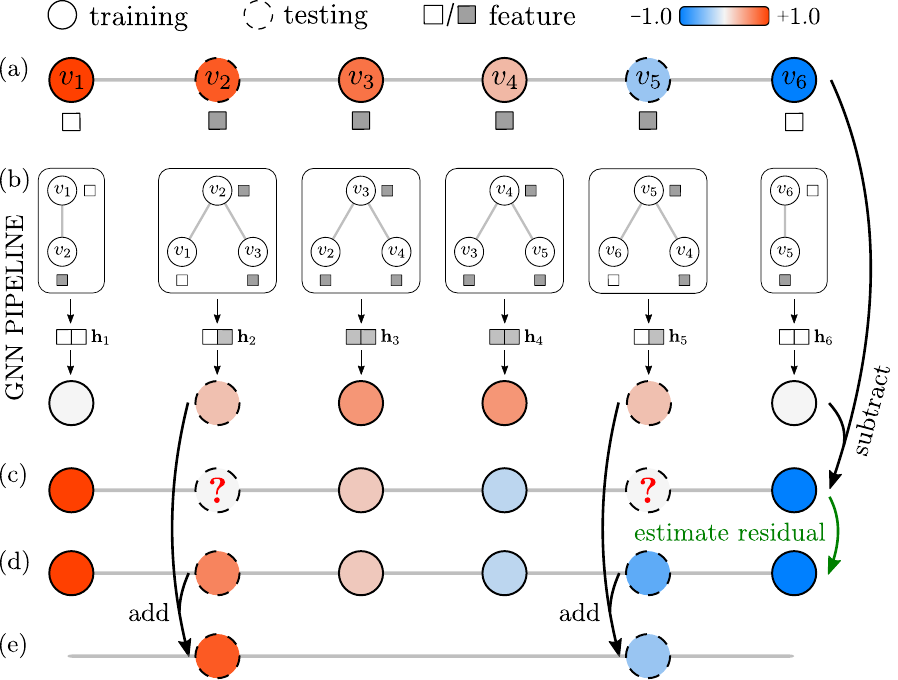}
\caption{Limitations of GNN regression and our proposed fix.
The vertex labels decrease from $v_{1}$ $(+1.0)$ to $v_{6}$ $(-1.0)$, and most interior vertices have positive labels.
(a) Each vertex's degree is used as its feature, and vertices are colored based on their labels.
The training vertices are $v_{1}, v_{3}, v_{4}, v_{6}$.
(b) The GNN encodes vertex neighborhoods by vectors $\mathbf{h}_{i}$, which are used independently for label prediction.
The GNN captures the positive trend for interior vertices but fails to distinguish $v_{1}, v_{2}, v_{3}$ from $v_{6}, v_{5}, v_{4}$ due to graph symmetry.
(c) GNN regression residuals for the training vertices.
(d) Our \emph{Correlated GNN} method estimates the residuals on testing vertices $v_{2}, v_{5}$.
(e) The estimated residuals are added to GNN outputs as our final predictions, yielding good estimates.
}
\label{fig:intro}
\end{figure}

\begin{figure*}[t]
\phantomsubfigure{fig:demo_a}
\phantomsubfigure{fig:demo_b}
\phantomsubfigure{fig:demo_c}
\phantomsubfigure{fig:demo_d}
\phantomsubfigure{fig:demo_e}
\phantomsubfigure{fig:demo_f}
\centering
\includegraphics[width=0.90\linewidth]{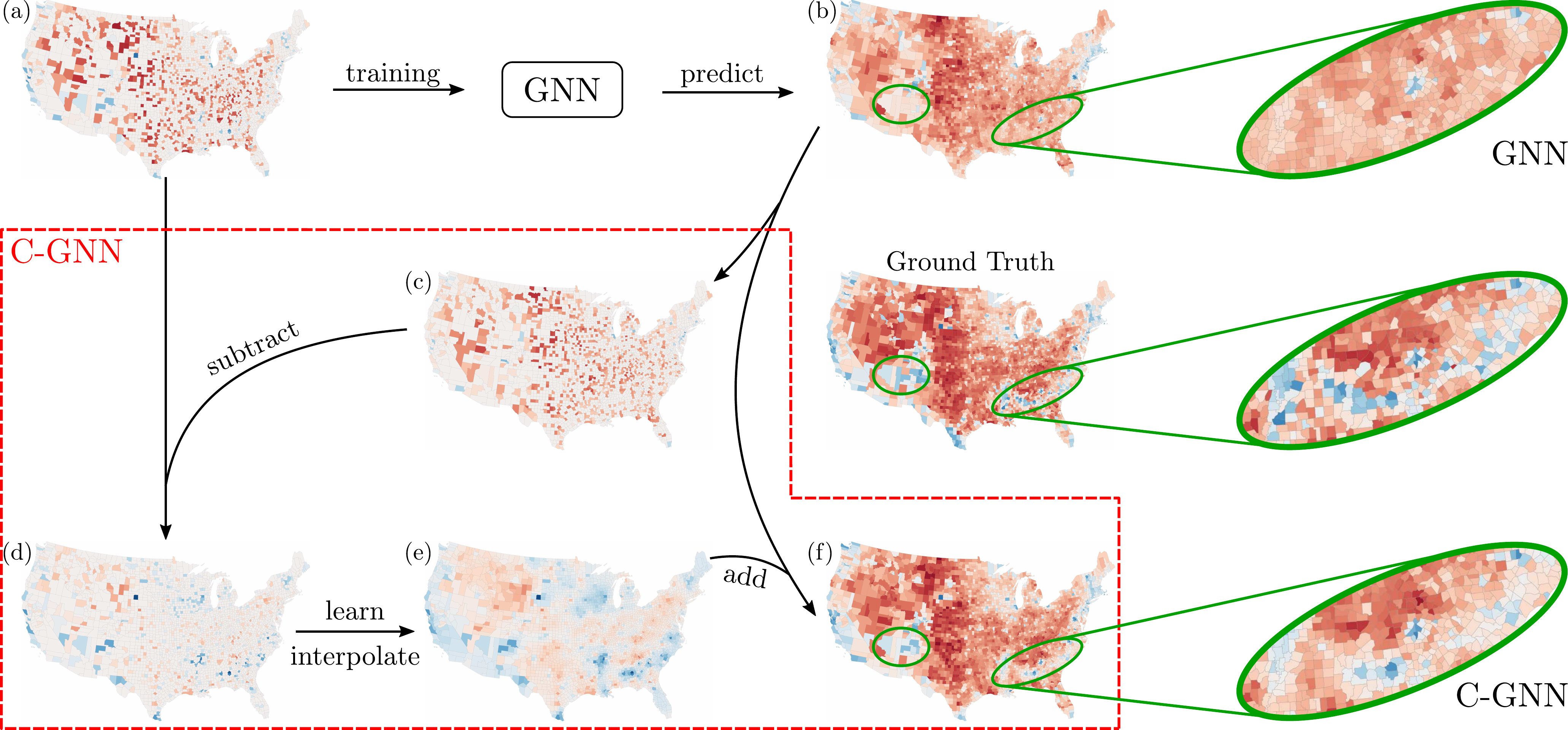}
\caption{Our Correlated GNN (C-GNN) framework for predicting county-level outcomes in the 2016 U.S.\ presidential election.
(a) The inputs are the county adjacency matrix, county-level demographic features, and 30\% of the labels.
(b) The GNN makes base predictions.
(c--d) The GNN predictions on the training data (c) show that the regression residual (d) is correlated amongst neighboring counties.
(e) Our C-GNN model learns the residual correlation and interpolates to get the residual on testing counties.
(f) Adding predicted residuals on the test counties to the GNN base prediction substantially increases accuracy.
}
\label{fig:demo}
\end{figure*}

These problems fall under the umbrella of semi-supervised learning for graph-structured data.
In the standard setup, one attribute (label) is observed only on a subset of vertices, and the goal is to predict missing labels 
using the graph topology, observed labels, and other observed attributes (features).
Graph neural networks (GNNs) are a class of methods that have had great success on such tasks~\cite{Kipf_2016,Hamilton_2017,Velickovi_2018,zhou2018graph},
largely due to their ability to extract information from vertex features.
%
%
The basic idea of GNNs is to first encode the local environment of each vertex into a vector
representation by transforming and aggregating its own features
along with the features of its neighbors in the graph~\cite{Hamilton-2017-representation},
with the label prediction at a node made from its vector representation.
Many target applications are for classification.\footnote{Perhaps the most well-studied problem
in this space is predicting the ``type'' of an academic paper in a citation network.}
In this paper, we focus on regression problems.
For example, in our U.S.\ election example above, a candidate might want to predict
their vote share in each county to plan a campaign strategy.
Existing GNN architectures can easily be adapted for regression
problems by simply changing the output layer and choosing a loss function such as the squared error in the
predicted value; automatic differentiation handles the rest.

However, a fundamental limitation of GNNs is that they predict each
vertex label independently given the set of vertex representation and ignore label
correlation of neighboring vertices.
Specifically, a prediction depends on the features of a vertex and other vertices in its neighborhood but not on the predictions
of neighboring vertices.
While not stated in these terms, similar observations have been made
about such limitations of GNNs~\cite{Qu_2019,Xu_2018,you2019position}.
\Cref{fig:intro_a,fig:intro_b} illustrates why this is problematic,
using a graph with topological and feature symmetry but monotonically varying vertex labels.
In this example, a GNN fails to distinguish vertex $v_{2}$ from $v_{4}$ and therefore cannot
predict correct labels for both of them.
On the other hand, traditional graph-based semi-supervised learning algorithms
(e.g., those based on label propagation~\cite{Zhu_2003,Zhou_2004}), work very well in
this case as the labels vary smoothly over the graph.
Of course, in many applications, vertex features are remarkably informative.
Still, gains in performance on benchmark tasks from using vertex features
have in some sense put blinders on the modeler --- the methodological focus is
on squeezing more and more information from the features~\cite{zhou2018graph},
ignoring signal in the joint distribution of the outcome.

In \cref{fig:intro}, vertex features partially explain the outcomes. The
features are somewhat --- but not overwhelmingly --- predictive. The question
then arises: when features are only somewhat predictive, can we get bigger gains
in predictive power by exploiting outcome correlations, rather than squeezing
minuscule additional signal in features with more complicated architectures?

\xhdr{The present work: Correlated Graph Neural Networks}
To answer the above question in the affirmative, we propose 
\emph{Correlated Graph Neural Networks} (C-GNNs).
The basic idea of C-GNNs is to use a GNN as a base regressor to capture the (possibly mild) outcome dependency on vertex features
and then further model the regression residuals on all vertices (\cref{fig:intro_c,fig:intro_d,fig:intro_e}).
While one can model the residual in many ways, we use a simple multivariate Gaussian
with a sparse precision matrix based on the graph topology.
At training, we learn the correlation structure by maximizing the marginal likelihood of the observed vertex labels.
At inference, we predict the outcomes on testing vertices by maximizing their probability conditioned on the training labels.
Importantly, our method covers the original GNN as a special case: minimizing
squared-error loss with respect to the GNN is the maximum likelihood estimator
when the precision matrix is the identity (errors are uncorrelated).
We also make no assumption on the GNN architecture, as our methodology ``sits on top''
of the base regressor.

For a real-world data example, we predict the county-level margin of victory in the 2016 U.S.\ presidential election (\cref{fig:demo}).
Each county is represented by a vertex with demographic features such as 
median household income, education levels, and unemployment rates,
and edges connect bordering counties.
While the GNN captures correlation between vertex features and outcomes
(\cref{fig:demo_a,fig:demo_b}), our C-GNN leverages residual correlation
(\cref{fig:demo_c,fig:demo_d,fig:demo_e,fig:demo_f}) to boost test
$R^{2}$ from $0.45$ to $0.63$.
The green circles show regions where the GNN produces large errors that are corrected by C-GNN.

More generally, we can replace the GNN base regressor with any feature-based predictor, e.g., a linear model or multilayer perceptron,
and our regression pipeline is unchanged.
With a linear model, for example, our framework is essentially performing generalized least squares~\cite{shalizi-2015-GLS},
where the precision matrix structure is given by the graph.
In practice, we find that within our framework, a GNN base regressor indeed works well for graph-structured data.

Our C-GNN consistently outperforms the base GNN and other competing baselines by large margins:
C-GNNs achieves a mean 14\% improvement in $R^{2}$ over GNNs for our datasets.
Furthermore, using a simple multilayer perceptron (that does not use neighborhood features) as the base regressor, our framework even outperforms a standard GNN in most experiments.
This highlights the importance of outcome correlation and suggests that focusing on minor GNN architecture improvements may not always be the right strategy.

Thus far, we have considered transductive learning, but
another standard setup for machine learning on graphs is \emph{inductive learning}: a
model is trained on one graph where labels are widely available and deployed on
other graphs where labels are more difficult to obtain.
Assuming that the learned GNN and the residual correlation generalize to
unseen graphs, our framework can simply condition on labeled vertices (if
available) in a new graph to improve regression accuracy.
Indeed, these assumptions hold for many real-world datasets that we consider.
With a small fraction of labels in the new graphs, inductive accuracies of
our C-GNN are even better than transductive accuracies of a GNN.
For example, we train a model to predict county-level unemployment rates using
60\% of labeled vertices in 2012.
Given 10\% of labels in the 2016 data, C-GNN achieves $0.65$ test $R^{2}$ on
unlabeled vertices, which is even more accurate than GNN trained directly on 60\% of
2016 labels ($R^{2} = 0.53$).

We also develop efficient numerical techniques that make model optimization tractable.
Standard factorization-based algorithms for the log marginal likelihood and derivative computations require $\mathcal{O}(n^{3})$ operations,
where $n$ is the number of nodes; such approaches do not scale beyond graphs with a few thousand vertices.
To remedy this, we use stochastic estimation~\cite{Ubaru_2017,Gardner_2018} to take full advantage of our
sparse and well-conditioned precision matrix, which reduces the computational
scaling to $\mathcal{O}(m)$, where $m$ is the number of edges, producing
low-variance unbiased estimates of model parameters.
We further introduce a simplified version of our method that assumes positive
correlation among neighboring residuals, which is common in real-world data.
The algorithm is extremely simple: train a standard GNN and then run label
propagation to interpolate GNN residuals on the testing vertices.
We call this LP-GNN and find that it also outperforms standard GNNs by a wide
margin on a variety of real-world datasets.


\section{Modeling residual correlation \label{sec:method}}
Let $G = (V, E, \{\mathbf{x}_{i}\})$ be a graph, where $V$ is the vertex set ($n = \lvert V \rvert$),
$E$ is the edge set ($m = \lvert E \rvert$),
and $\mathbf{x}_{i}$ denotes the features for vertex $i \in V$.
We consider the semi-supervised vertex label regression problem: given real-valued
labels\footnote{Since we are performing regression, ``labels'' might also be called ``outcomes'', or ``targets'';
we use those terms interchangeably in this paper.}
$y_L$ on a subset of vertices $L \subseteq V$, predict labels on the rest of the vertices $U \equiv V \backslash L$.
In this section, we first review GNNs and discuss its implicit statistical assumptions.
As we show in \cref{sec:experiments}, these assumption are often invalid for real-world graph data.
Motivated by this insight, we improve the predictive power of GNNs by explicitly
modeling label correlations with a multivariate Gaussian, and
introduce efficient numerical techniques for learning model parameters.

\subsection{Statistical Interpretation of Standard GNNs}
In a standard GNN regression pipeline, the features in the neighborhood of a vertex get encoded into a vertex representation,%
\footnote{%
  For instance, a $K$-step graph convolution network (GCN) computes vertex representations by repeated local feature averaging, transformation, and nonlinear activation:
\[\mathbf{h}_{i}^{(0)} = \mathbf{x}_{i}; \hspace{0.05in} \mathbf{h}_{i}^{(k)} = \phi\left(\mathbf{W}^{(k)} \cdot \textsc{mean}\left(\{\mathbf{h}_{i}^{(k-1)}\} \cup \{\mathbf{h}_{j}^{(k-1)}: j \in N_{1}(i)\}\right)\right); \hspace{0.05in} \mathbf{h}_{i} = \mathbf{h}_{i}^{(K)},\]
where $\mathbf{W}^{(k)}$ is a weight matrix at step $k$, and $\phi$ is a nonlinear activation function.
}
and each vertex representation is used independently for label prediction:
\begin{align}
\mathbf{h}_{i} = f\left(\mathbf{x}_{i}, \{\mathbf{x}_{j}: j \in N_{K}(i)\}, \theta\right); \qquad \hat{y}_{i} = g(\mathbf{h}_{i}, \theta).
\label{eq:gnn}
\end{align}
Here, $N_{K}(i)$ denotes the $K$-hop neighborhood of vertex $i$.
Oftentimes, $K = 2$~\cite{Kipf_2016,Hamilton_2017}.
The GNN weights $\theta$ are trained using observed labels, and the most common loss for regression is the squared error:
\begin{align}
\textstyle \sum_{i \in L} (g(\mathbf{h}_{i}, \theta) - y_{i})^2.
  \label{eq:gnn_loss}
\end{align}

Following statistical arguments for ordinary least squares~\cite{friedman2001elements}, minimizing \cref{eq:gnn_loss}
is equivalent to maximizing the likelihood of a factorizable joint distribution of labels, where the distribution
of each label conditioned on the vertex representation is a univariate Gaussian:
\begin{align}
p(\mathbf{y} \given G) = \prod_{i \in V} p(y_{i} \given \mathbf{h}_{i}); \quad y_{i} \given \mathbf{h}_{i} \sim \mathcal{N}(\hat{y}_{i}, \sigma^{2})
\end{align}
Consequently, the errors in the estimates $y_i - \hat{y}_i$ are i.i.d.\ with mean zero.
However, there's no reason to assume independence, and in cases such as election data, accounting
for error correlation is critical.\footnote{\url{https://fivethirtyeight.com/features/a-users-guide-to-fivethirtyeights-2016-general-election-forecast/}}
We thus consider correlation structure next.

\subsection{Correlation as a Multivariate Gaussian \label{subsec:precision_parametrization}}
We model the joint distribution of labels as a multivariate Gaussian:
\begin{align}
    \mathbf{y} \sim \mathcal{N}\left(\hat{\mathbf{y}}, \Gamma^{-1}\right) \hspace{0.05in} \textrm{or equivalently}, \hspace{0.05in} \mathbf{r} \equiv \mathbf{y} - \hat{\mathbf{y}} \sim \mathcal{N}\left(0, \Gamma^{-1}\right),
    \label{eq:mg}
\end{align}
where $\Gamma = \Sigma^{-1}$ is the inverse covariance (or precision) matrix, and $\mathbf{r}$ is the residual of GNN regression.
Here, we parameterize the precision matrix in a way that (i) uses the graph topology and (ii) will be computationally tractable:
\begin{align}
    \Gamma = \beta (\mathbf{I} - \alpha \mathbf{S}),
    \label{eq:precision}
\end{align}
where $\mathbf{I}$ is the identity matrix and $\mathbf{S} = \mathbf{D}^{-1/2} \mathbf{A} \mathbf{D}^{-1/2}$ is the normalized adjacency matrix.
The scalar $\beta$ controls the overall magnitude of the residual, and
the scalar $\alpha$ reflects the correlation structure.
The sign of $\alpha$ is the direction of correlation (positive or negative), and the magnitude measures the strength of correlation.

Validity of the multivariate Gaussian requires that $\Gamma$ is positive definite.
This requirement is easily satisfied by restricting $-1 < \alpha < 1$ and $\beta > 0$. 
First, we verify both $(\mathbf{I} + \mathbf{S})$ and $(\mathbf{I} - \mathbf{S})$ are positive semi-definite by expanding their quadratic form with any $\mathbf{z} \in \mathbb{R}^{n}$:
\begin{align}
\textstyle    \mathbf{z}^{\intercal} (\mathbf{I} + \mathbf{S}) \mathbf{z} &= \textstyle \sum_{(i,j) \in E} \left(\nicefrac{z_{i}}{\sqrt{D_{ii}}} + \nicefrac{z_{j}}{\sqrt{D_{jj}}}\right)^{2} \geq 0 \\
\textstyle    \mathbf{z}^{\intercal} (\mathbf{I} - \mathbf{S}) \mathbf{z} &= \textstyle \sum_{(i,j) \in E} \left(\nicefrac{z_{i}}{\sqrt{D_{ii}}} - \nicefrac{z_{j}}{\sqrt{D_{jj}}}\right)^{2} \geq 0
\end{align}
For $0 \le \alpha < 1$, $\Gamma = (1 - \alpha) \beta \mathbf{I} + \alpha \beta (\mathbf{I} - \mathbf{S}) \succ 0$ since the first term is strictly positive definite, and the second term is positive semi-definite.
A similarly argument holds for $-1 < \alpha < 0$.
Two special cases of the precision matrix in \cref{eq:precision} deserve special attention.
First, when $\alpha = 0$, $\Gamma$ is the identity matrix (up to constant scaling), and the model reduces to standard GNN regression.
Second, in the limit $\alpha \rightarrow 1$, $\Gamma$ is the normalized Laplacian matrix,
and the noise is assumed to be smooth over the entire graph. The normalized
Laplacian matrix is only positive semi-definite, so we make sure the limit is never realized in practice;
however, we use this as motivation for a simplified version of the model in
\cref{subsec:lp_gnn}.

\xhdr{Inferring unknown labels}
\begin{algorithm}[t]
\SetKwInOut{Input}{Input}
\SetKwInOut{Output}{Output}
\Input{\ normalized adjacency matrix $\mathbf{S}$; features $\{\mathbf{x}_{i}\}$; training labels $\mathbf{y}_{L}$; parameters $\alpha, \beta$; GNN weights $\theta$}
\Output{\ predicted labels $\mathbf{y}_{U}^{\rm C-GNN}$ for unknown vertices}
\BlankLine
$\Gamma \gets \beta (\mathbf{I} - \alpha \mathbf{S})$ \Comment{precision matrix} \\
$\mathbf{h}_{i} \gets f(\mathbf{x}_{i}, \{\mathbf{x}_{j}: j \in N_{K}(i)\}, \theta),\ \forall i \in V$ \Comment{GNN learning} \\
$\hat{y}_{i} \gets g(\mathbf{h}_{i}, \theta),\ \forall i \in V$ \Comment{GNN predictions} \\
$\mathbf{r}_{L} \gets \mathbf{y}_{L} - \hat{\mathbf{y}}_{L}$ \Comment{training residuals} \\
$\mathbf{y}_{U}^{\rm C-GNN} \gets \hat{\mathbf{y}}_{U} - \Gamma_{UU}^{-1} \Gamma_{UL} \mathbf{r}_{L}$ \Comment{C-GNN predictions} \\
\caption{C-GNN label inference.}
\label{alg:inference}
\end{algorithm}
Now we show how to infer unlabeled vertices assuming $\alpha$, $\beta$, $\theta$, and $y_L$ are known.
If we partition \cref{eq:mg} into the labeled and unlabeled blocks,
\begin{align}
    \begin{bmatrix}
    \mathbf{y}_{L} \\
    \mathbf{y}_{U}
    \end{bmatrix}
    \sim
    \mathcal{N}\left(
    \begin{bmatrix}
    \hat{\mathbf{y}}_{L} \\
    \hat{\mathbf{y}}_{U}
    \end{bmatrix},
    \begin{bmatrix}
    \Gamma_{LL} & \Gamma_{LU} \\
    \Gamma_{UL} & \Gamma_{UU}
    \end{bmatrix}^{-1}
    \right),
    \label{eq:mg_partition}
\end{align}
then conditioned on the labeled vertices $L$, the distribution of vertex labels on $U$ is also a multivariate Gaussian,
\begin{align}
    \mathbf{y}_{U} \given \mathbf{y}_{L} \sim \mathcal{N}\left(\hat{\mathbf{y}}_{U} - \Gamma_{UU}^{-1} \Gamma_{UL} \mathbf{r}_{L}, \Gamma_{UU}^{-1}\right).
\end{align}
Our model uses the expectation of this conditional distribution as the final prediction, which is given by the Gaussian mean,
\begin{align}
    \mathbf{y}_{U}^{\rm C-GNN} = \hat{\mathbf{y}}_{U} - \Gamma_{UU}^{-1} \Gamma_{UL} \mathbf{r}_{L}.
\end{align}
\Cref{alg:inference} summarizes the inference algorithm.
Next, we consider learning optimal parameters from labeled data.

\xhdr{Learning model parameters}
\begin{algorithm}[t]
\SetKwInOut{Input}{Input}
\SetKwInOut{Output}{Output}
\Input{\ normalized adjacency matrix $\mathbf{S}$; features $\{\mathbf{x}_{i}\}$; all training vertices $L_{0}$, labels $\mathbf{y}_{L_{0}}$; number of training steps $p$; batch size $b$}
\Output{\ optimized $\alpha, \beta, \theta$}
\BlankLine
randomly initialize $\alpha, \beta, \theta$ \\
\For{$i \gets 1$ \rm{to} $p$}{
    $\Gamma \gets \beta (\mathbf{I} - \alpha \mathbf{S})$ \\
    $L \gets$ \texttt{subsample}($L_{0}$, $b$) \Comment{get mini-batch} \\
    $\mathbf{h}_{i} \gets f(\mathbf{x}_{i}, \{\mathbf{x}_{j}: j \in N_{K}(i)\}, \theta),\ \forall i \in L$ \\
    $\hat{y}_{i} \gets g(\mathbf{h}_{i}, \theta),\ \forall i \in L$ \Comment{GNN predictions} \\
    $\mathbf{r}_{L} \gets \mathbf{y}_{L} - \hat{\mathbf{y}}_{L}$ \Comment{training residuals} \\
    $\Omega \gets \mathbf{r}_{L}^{\intercal} \bar{\Gamma}_{LL} \mathbf{r}_{L} - \log\det(\Gamma) + \log\det(\Gamma_{UU})$ \\
    compute $\nicefrac{\partial \Omega}{\partial \alpha}, \nicefrac{\partial \Omega}{\partial \beta}, \nicefrac{\partial \Omega}{\partial \theta}$ \Comment{\cref{eq:loss_derivatives}} \\
    $\alpha, \beta, \theta \gets$ \texttt{update}($\Omega, \nicefrac{\partial \Omega}{\partial \alpha}, \nicefrac{\partial \Omega}{\partial \beta}, \nicefrac{\partial \Omega}{\partial \theta}$) \\
}
\caption{C-GNN training (mini-batched).}
\label{alg:trainings}
\end{algorithm}
Given the observed outcomes $y_L$ on $L$, the precision matrix parameters $\alpha, \beta$
and GNN weights $\theta$ are learned by maximum likelihood estimation.
The marginal distribution of the GNN residual on $L$ is a multivariate Gaussian~\cite{rasmussen2003gaussian}:
\begin{align}
    \mathbf{r}_{L} = \mathbf{y}_{L} - \hat{\mathbf{y}}_{L} \sim \mathcal{N}\left(0, \bar{\Gamma}_{LL}^{-1}\right),
\end{align}
where $\bar{\Gamma}_{LL} = \Gamma_{LL} - \Gamma_{LU} \Gamma_{UU}^{-1} \Gamma_{UL}$ is the corresponding precision matrix.
We define the loss function as the negative log marginal likelihood of observed labels:
\begin{align}
    \Omega &= -\log p(\mathbf{y}_{L}|\alpha, \beta, \theta) \nonumber \\
           &= \left[\mathbf{r}_{L}^{\intercal} \bar{\Gamma}_{LL} \mathbf{r}_{L} - \log\det(\bar{\Gamma}_{LL}) + n \log(2\pi)\right] / 2 \nonumber \\
           &\propto \hspace{0.05in} \mathbf{r}_{L}^{\intercal} \bar{\Gamma}_{LL} \mathbf{r}_{L} - \log\det(\Gamma) + \log\det(\Gamma_{UU})
    \label{eq:loss}
\end{align}
Then, the loss function derivatives with respect to the model parameters take the following expression,
\begin{align}
    \frac{\partial \Omega}{\partial \alpha} &= \mathbf{r}_{L}^{\intercal} \frac{\partial \bar{\Gamma}_{LL}}{\partial \alpha} \mathbf{r}_{L} - \texttt{tr}\left(\Gamma^{-1} \frac{\partial \Gamma}{\partial \alpha}\right) + \texttt{tr}\left(\Gamma_{UU}^{-1} \frac{\partial \Gamma_{UU}}{\partial \alpha}\right) \nonumber \\
    \frac{\partial \Omega}{\partial \beta} &= \mathbf{r}_{L}^{\intercal} \frac{\partial \bar{\Gamma}_{LL}}{\partial \beta} \mathbf{r}_{L} - \texttt{tr}\left(\Gamma^{-1} \frac{\partial \Gamma}{\partial \beta}\right) + \texttt{tr}\left(\Gamma_{UU}^{-1} \frac{\partial \Gamma_{UU}}{\partial \beta}\right) \nonumber \\
    \frac{\partial \Omega}{\partial \theta} &= -2 \mathbf{r}_{L}^{\intercal} \bar{\Gamma}_{LL} \frac{\partial \hat{\mathbf{y}}_{L}}{\partial \theta},
    \label{eq:loss_derivatives}
\end{align}
where $\nicefrac{\partial \hat{\mathbf{y}}_{L}}{\partial \theta}$ can be computed with back-propagation, and
\begin{align}
    \frac{\partial \bar{\Gamma}_{LL}}{\partial \alpha} = \frac{\partial \Gamma_{LL}}{\partial \alpha} &- \frac{\partial \Gamma_{LU}}{\partial \alpha} \Gamma_{UU}^{-1} \Gamma_{UL} + \Gamma_{LU} \Gamma_{UU}^{-1} \frac{\partial \Gamma_{UU}}{\partial \alpha} \Gamma_{UU}^{-1} \Gamma_{UL} \nonumber \\
    &- \Gamma_{LU} \Gamma_{UU}^{-1} \frac{\partial \Gamma_{UL}}{\partial \alpha} \nonumber \\
    \frac{\partial \bar{\Gamma}_{LL}}{\partial \beta} = \frac{\partial \Gamma_{LL}}{\partial \beta} &- \frac{\partial \Gamma_{LU}}{\partial \beta} \Gamma_{UU}^{-1} \Gamma_{UL} + \Gamma_{LU} \Gamma_{UU}^{-1} \frac{\partial \Gamma_{UU}}{\partial \beta} \Gamma_{UU}^{-1} \Gamma_{UL} \nonumber \\
    &- \Gamma_{LU} \Gamma_{UU}^{-1} \frac{\partial \Gamma_{UL}}{\partial \beta}.
    \label{eq:gamma_derivatives}
\end{align}
Finally, let $P,Q$ denote two arbitrary sets of vertices.
The derivatives of each precision matrix block $\Gamma_{PQ}$ are given by
$\nicefrac{\partial \Gamma_{PQ}}{\partial \alpha} = -\beta \mathbf{S}_{PQ}$ and $\nicefrac{\partial \Gamma_{PQ}}{\partial \beta} = \nicefrac{\Gamma_{PQ}}{\beta}$.
In practice, we employ a mini-batch sampling
for better memory efficiency, and we maximize the marginal likelihood of a
mini-batch at each training step (\cref{alg:trainings}).

One remaining issue is the computational cost.
Standard matrix factorization-based algorithms for computing the matrix inverse and log
determinant have complexity cubic in the number of vertices, which is
computationally prohibitive for graphs beyond a few thousand vertices.
In \cref{sec:efficient_optimization}, we show how to reduce these computations to linear
in the number of edges, using recent tricks in stochastic trace estimation.
Next, we offer an even cheaper alternative that works well when $\alpha$ is close to $1$.

\subsection{A Simple Propagation-based Algorithm}\label{subsec:lp_gnn}
\begin{algorithm}[t]
\SetKwInOut{Input}{Input}
\SetKwInOut{Output}{Output}
\Input{\ normalized adjacency matrix $\mathbf{S}$; features $\{\mathbf{x}_{i}\}$; training labels $\mathbf{y}_{L}$}
\Output{\ predicted labels $\mathbf{y}_{U}^{\rm LP-GNN}$ for unknown vertices}
\BlankLine
train standard GNN, get optimized parameter $\theta$ \\
$\mathbf{h}_{i} \gets f(\mathbf{x}_{i}, \{\mathbf{x}_{j}: j \in N_{K}(i)\}, \theta),\ \forall i \in V$ \\
$\hat{y}_{i} \gets g(\mathbf{h}_{i}, \theta),\ \forall i \in V$ \Comment{GNN predictions} \\
$\mathbf{r}_{L} \gets \mathbf{y}_{L} - \hat{\mathbf{y}}_{L}$ \Comment{training residuals} \\
$\mathbf{r}_{U}^{\rm est} \gets \texttt{LabelPropagation}(\mathbf{S}, \mathbf{r}_{L})$ \Comment{e.g., \cref{alg:lp}} \\
$\mathbf{y}_{U}^{\rm LP-GNN} \gets \hat{\mathbf{y}}_{U} + \mathbf{r}_{U}^{\rm est}$ \\
\caption{LP-GNN regression.}
\label{alg:lp_gnn}
\end{algorithm}
Our framework is inspired in part by label propagation~\cite{Zhu_2003,Zhou_2004}, where the neighboring correlation is always assumed to be positive.
In fact, if we fix $\alpha = 1$ and replace the base GNN regressor with one that
gives uniform 0 prediction for all vertices, our method reduces to a variant of
label propagation that uses the normalized Laplacian matrix (see details in
\cref{subsec:label_propagation}), which might be expected given the
connection between Gaussian Process regression (kriging) and
graph-based semi-supervised learning~\cite{Xu_2010}.

This observation motivates an extremely simple version of our method, which we call LP-GNN (\cref{alg:lp_gnn}):
(i) train a standard GNN;
(ii) run label propagation from the residuals on labeled vertices;
(iii) add the propagated result to the GNN predictions.
LP-GNN is a lightweight framework for data where residual correlation is strong
and positive, and in principle, any label propagation method could be employed.
We show in \cref{sec:experiments} that this provides substantial
improvements over a GNN in many cases, but the C-GNN still has better predictions.

\subsection{Extension to Multiple Edge Types \label{subsec:extended_model}}
Our model can also be extend to study graphs with multiple edge types.
For instance, later in \cref{subsec:transductive_results}, we consider a road traffic
network where different pairs of lanes, based on their orientations, are connected
with different types of edges.
In this setting, we decompose the total adjacency matrix as $\mathbf{A} = \sum_{i} \mathbf{A}^{(i)}$, where $\mathbf{A}^{(i)}$ is given by the edges of type $i$.
Then, denoting $\mathbf{S}^{(i)} = \mathbf{D}^{-1/2} \mathbf{A}^{(i)} \mathbf{D}^{-1/2}$, we parametrize the precision matrix as
\begin{align}
    \textstyle \Gamma = \beta(\mathbf{I} - \sum_{i} \alpha_{i} \mathbf{S}^{(i)}).
\end{align}
Following the same logic as in \cref{subsec:precision_parametrization}, the
above precision matrix is still positive definite if $-1 < \alpha_i < 1$ for all
$i$, and the loss function derivatives with respect to $\{\alpha_{i}\}$ are
similar to the original model.
The extended model provides finer grained descriptions for interactions among
neighboring vertices.
Our experiments show that the extended model captures the difference in correlation
strengths for different types of edges in the traffic network, as well as
improving the regression accuracy.


\section{Fast Model Optimization \label{sec:efficient_optimization}}
We have introduced a simple and interpretable framework to exploit residual correlations.
However, the model's applicability to large-scale networks is limited by the cubic-scaling cost
associated with the log-determinant computations during learning.
%
%
Here, we use stochastic estimation of the log determinant and its
derivatives. By taking advantage of our sparse precision
matrix parametrization, this makes computations
essentially linear in the size of the graph.
\subsection{Efficient Log-determinant Estimation \label{subsec:log_det_derivatives}}
The major computational cost in our framework boils down to three types of matrix computations:
(i) solving the linear system $\Gamma^{-1} \mathbf{z}$;
(ii) the matrix trace $\texttt{tr}(\Gamma^{-1} \frac{\partial \Gamma}{\partial \alpha})$; and
(iii) the log determinant $\log\det(\Gamma)$.%
\footnote{We focus on evaluating $\log\det(\Gamma)$ and $\rfrac{\partial \Omega}{\partial \alpha}$ in our analysis, but the results easily generalize to $\log\det(\Gamma_{UU})$ and $\rfrac{\partial \Omega}{\partial \beta}$.}
Next, we show how our precision matrix parametrization allows those operations to be computed efficiently using
conjugate gradients (CG), stochastic trace estimation, and Lanczos quadrature~\cite{Avron_2011,Fitzsimons_2018,Ubaru_2017,dong2019network}.

\xhdr{Conjugate Gradients (CG) solution of $\Gamma^{-1} \mathbf{z}$}
CG is an iterative algorithm for solving linear systems when the matrix is symmetric positive definite.
Each CG iteration computes one matrix vector multiplication and a handful of
vector operations, so approximately solving $\Gamma^{-1} \mathbf{z}$ with $k$
CG iterations requires $\mathcal{O}(k m)$ operations, where $m$ is the number
of edges in the graph.
The convergence rate of CG depends on the condition number of $\Gamma$, 
which is the ratio between the largest and smallest eigenvalues: $\kappa(\Gamma) = \lambda_{\rm max}(\Gamma) / \lambda_{\rm min}(\Gamma)$.
In particular, for a fixed error tolerance, CG converges in $\mathcal{O}(\sqrt{\kappa(\Gamma)})$ iterations.
We now provide an upper bound on $\kappa(\Gamma)$, which justifies using a fixed number of iterations.

Since the eigenvalues of the normalized adjacency matrix $\mathbf{S}$ are bounded between $-1.0$ and $1.0$~\cite{chung1997spectral},
we can bound the extreme eigenvalues of the precision matrix as follows:
\begin{align}
    \lambda_{\rm max}(\Gamma) &= \beta \lambda_{\rm max}(\mathbf{I} - \alpha \mathbf{S}) < \beta [\lambda_{\rm max}(\mathbf{I}) + \lambda_{\rm max}(-\alpha \mathbf{S})] = \beta(1 + |\alpha|) \nonumber \\
    \lambda_{\rm min}(\Gamma) &= \beta \lambda_{\rm min}\left[(1 - |\alpha|) \mathbf{I} + |\alpha| \left(\mathbf{I} - \frac{\alpha}{|\alpha|}\mathbf{S}\right)\right] > \beta(1 - |\alpha|)
\end{align}
Then, the upper bound of the condition number is
\begin{align}
    \kappa(\Gamma) = \lambda_{\rm max}(\Gamma) / \lambda_{\rm min}(\Gamma) < (1+|\alpha|) / (1-|\alpha|),
\end{align}
which does not depend on the graph topology.
(This upper bound also applies to $\Gamma_{UU}$ via the eigenvalue interlacing theorem.)
Therefore, by further constraining $|\alpha| < 1 - \eta$ for a small positive constant $\eta$, CG algorithm converges in $\mathcal{O}(\sqrt{2/\eta})$ iterations.
We will verify numerically in \cref{sec:approximation_validation} that in practice,
CG converges in a couple dozen iterations for our framework.

\xhdr{Stochastic Estimation of $\texttt{tr}(\Gamma^{-1} \frac{\partial \Gamma}{\partial \alpha})$ }
The stochastic trace estimator is an established method for approximating the trace of a matrix function~\cite{Hutchinson_1989,Avron_2011,Ubaru_2017}.
Given a Gaussian random vector $\mathbf{z} \sim \mathcal{N}(0, \mathbf{I})$ with $E[z_{i} z_{j}] = \delta_{ij}$, where $\delta_{ij}$ is the Kronecker delta function,
\begin{align}
\textstyle    \mathbb{E}\left[\mathbf{z}^{\intercal} \mathbf{M} \mathbf{z}\right] = \mathbb{E}\left[\sum_{i} z_{i}^{2} M_{ii} + \sum_{i \neq j} z_{i} z_{j} M_{ij}\right] = \sum_{i} M_{ii}
\end{align}
gives the unbiased trace estimation for any matrix $\mathbf{M}$.
This allows us to estimate $\texttt{tr}(\Gamma^{-1} \frac{\partial \Gamma}{\partial \alpha})$ without explicitly forming $\Gamma^{-1}$.
In practice, given $T$ independent and identically sampled Gaussian random vectors 
$\mathbf{z}_{t} \sim \mathcal{N}(0, \mathbf{I})$, $t = 1, \ldots, T$,
we estimate the matrix trace by
\begin{align}
\textstyle    \texttt{tr}(\Gamma^{-1} \frac{\partial \Gamma}{\partial \alpha}) = \mathbb{E}\left[\mathbf{z}_{t}^{\intercal} \Gamma^{-1} \frac{\partial \Gamma}{\partial \alpha} \mathbf{z}_{t}\right] \approx \frac{1}{T} \sum_{t=1}^{T} \left(\Gamma^{-1} \mathbf{z}_{t}\right)^{\intercal} \left(\frac{\partial \Gamma}{\partial \alpha} \mathbf{z}_{t}\right),
\end{align}
which would require calling the conjugate gradient solver $T$ times with the same matrix $\Gamma$ but different right-hand-sides.

\xhdr{Stochastic Lanczos quadrature for $\log\det(\Gamma)$}
We adopt the approach of Ubaru et al. for approximating the log-determinant, which estimates the trace of the logarithm of the matrix~\cite{Ubaru_2017}:
\begin{align}
    \log\det(\Gamma) = \texttt{tr}(\log \Gamma) &\approx \textstyle \frac{1}{T} \sum_{t=1}^{T} \mathbf{z}_{t}^{\intercal} \log \Gamma \mathbf{z}_{t} \nonumber \\
    &= \textstyle \frac{1}{T} \sum_{t=1}^{T} \mathbf{z}_{t}^{\intercal} \mathbf{Q} \log \Lambda \mathbf{Q}^{\intercal} \mathbf{z}_{t} \nonumber \\
    &= \textstyle \frac{1}{T} \sum_{t=1}^{T} \sum_{i=1}^{n} \mu_{ti}^{2} \cdot \log \lambda_{i}(\Gamma),
\end{align}
where $\Gamma = \mathbf{Q} \Lambda \mathbf{Q}^{\intercal}$ is the eigen-decomposition, and $\mu_{ti}$ is the projected length of $\mathbf{z}_{t}$ on the $i$-th eigenvector of $\Gamma$.
The expression $\sum_{i}^{n} \mu_{ti}^{2} \cdot \log \lambda_{i}(\Gamma)$ can be considered as a Riemann-Stieltjes integral, and is further approximated with Gaussian quadrature:
\begin{align}
\textstyle    \sum_{i=1}^{n} \mu_{ti}^{2} \cdot \log \lambda_{i}(\Gamma) \approx \sum_{i=1}^{k} w_{ti}^{2} \cdot \log \xi_{ti},
\end{align}
where the optimal nodes $\{\xi_{ti}\}$ and weights $\{w_{ti}\}$ for numerical integration are determined as follows.
First, run $k$ steps of the Lanczos algorithm with $\Gamma$ and initial vector $\mathbf{z}_{t}$ to get $\mathbf{V}_{t}^{\intercal} \Gamma \mathbf{V}_{t} = \mathbf{T}_{t}$. 
Then, perform the eigen-decomposition of the tri-diagonal matrix $\mathbf{T}_{t} = \mathbf{P}_{t} \Xi_{t} \mathbf{P}_{t}^{\intercal}$. 
Each integration node is an eigenvalue of $\mathbf{T}_{t}$ whose weight is the first element of each corresponding eigenvector:
\begin{align}
    \xi_{ti} = (\Xi_{t})_{ii}, \qquad w_{ti} = \sqrt{n} \cdot (\mathbf{P}_{t})_{1i}
\end{align}
Please see Ubaru et al.\ for a complete derivation~\cite{Ubaru_2017}.

\xhdr{Implementation and algorithm complexity}
Both the CG and Lanczos algorithms are Krylov subspace methods, and their convergence essentially depends on the condition number~\cite{Watkins2007}.
Since the condition number in our precision matrix parametrization is bounded, we use a fixed number of $k$ iterations in both algorithms.
Furthermore, the error of the stochastic trace estimator decrease with the
number of trial vectors $T$ as $\mathbf{O}(T^{-1/2})$, regardless
of the graph topology, and we also use a fixed number of $T$ vectors.

We summarize the overall complexity of the proposed method for evaluating \cref{eq:loss,eq:loss_derivatives} in each optimization step.
Computing $\hat{\mathbf{y}}_{L}$ and $\rfrac{\partial \hat{\mathbf{y}}_{L}}{\partial \theta}$ through forward and back propagation takes $\mathbf{O}(n)$ operations
(assuming constant-size neighborhood subsampling in the GNN implementation).
Evaluating the quadratic forms in \cref{eq:loss_derivatives} invokes a constant number of calls ($8$ in our case) to the CG solver, which takes $\mathbf{O}(mk)$ operations.
The trace estimations $\texttt{tr}(\Gamma^{-1} \frac{\partial \Gamma}{\partial \alpha})$ invokes $T$ calls to the CG solver, which takes $\mathbf{O}(mkT)$ operations.
The log-determinant estimation $\log\det(\Gamma)$ invokes $T$ calls to the Lanczos algorithm, which takes $\mathbf{O}(mkT)$ operations.
Finally, the eigen-decomposition of the $k$-by-$k$ tri-diagonal matrices $\{\mathbf{T}_{t}\}_{t=1}^{T}$ takes $\mathcal{O}(T k^{2})$ operations.
We choose $T=128, k=32$ as the default hyperparameters, independent of the size of the graph for an overall complexity of $\mathcal{O}(m)$,
i.e., linear in the number of edges.

Stochastic estimation of the log determinant and its derivatives of the covariance matrix has been considered in the context of Gaussian Processes~\cite{Gardner_2018}, where a similar computational scheme is used to reduce the complexity from $\mathcal{O}(n^{3})$ to $\mathcal{O}(n^{2})$.
Our model further benefits from the sparse and well-conditioned precision matrix parametrization, which results in linear-time computations of the objective function and its gradients.
We implement the log-determinant estimation function in Julia using the \texttt{Flux.jl} automatic differentiation package~\cite{Innes_2018},
which automatically tracks the function value and derivatives (see \cref{subsec:flux_implementation} for details).
We also adapt techniques proposed by Gardner et al.\ for reusing computation and improving cache efficiency~\cite{Gardner_2018}.

\subsection{Validation of Stochastic Estimation \label{sec:approximation_validation}}
%
%
We now demonstrate the accuracy of the proposed stochastic estimation algorithm as a function of the hyperparameters $T$ and $k$.
We find that the proposed scheme gives accurate and unbiased estimates for the log determinant and its derivatives for modest
values of $T$ and $k$, and we empirically show linear scaling.

\xhdr{Accuracy in estimating log determinant and its derivatives}
\begin{figure}[t]
\centering
\includegraphics[width=1.00\linewidth]{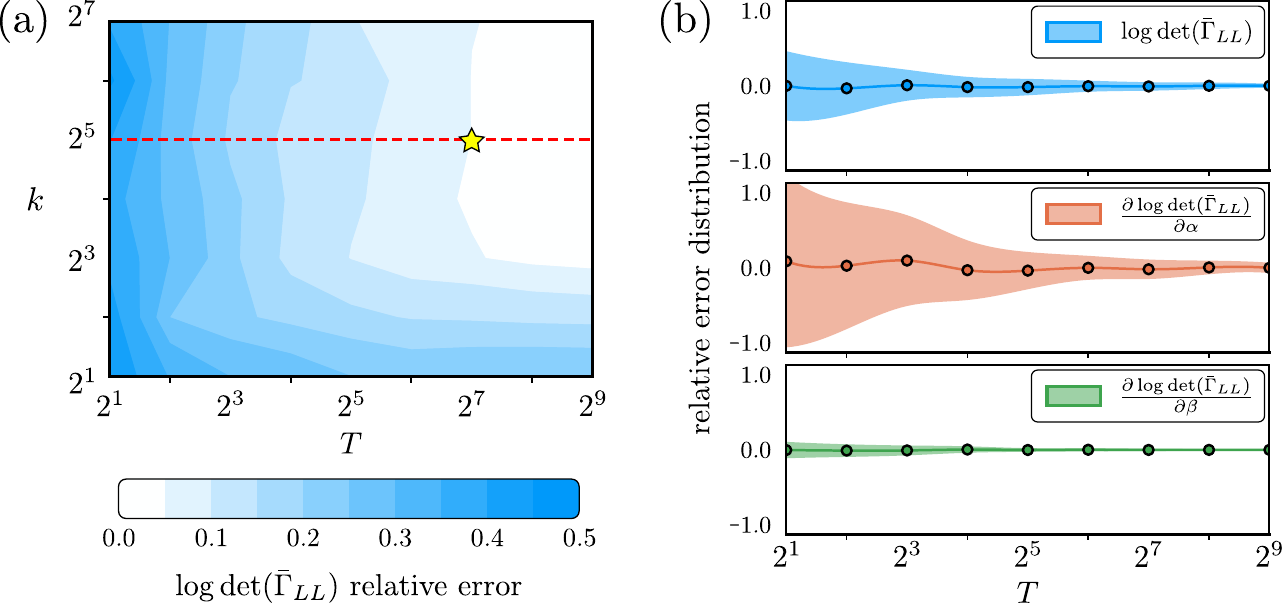}
\caption{Estimation error as a function of hyperparameters.
(a) Relative error of log determinant estimation. The yellow star marks our default hyperparameters.
(b) Relative error distribution along the red dashed line in (a) of the log determinant and its derivatives as a function of $T$.
}
\label{fig:accuracy}
\end{figure}
To test our fast algorithms, we sample a Watts-Strogatz graph~\cite{watts1998collective} with $500$ vertices and average vertex degree $10$.
We randomly select $50\%$ vertices as labeled, and compute the marginal precision matrix $\bar{\Gamma}_{LL}$ with $\alpha = 0.999$ and $\beta = 1.0$,
which corresponds to an ill-conditioned parametrization.
To understand how the quality of the approximation depends on the
hyperparameters, we compare our stochastic algorithm output to ``ground truth'' log-determinant and derivatives obtained from
factorization-based algorithms.
The estimation accuracy is measured by the root mean square relative error over $100$ runs (for various $T,k$; \cref{fig:accuracy}).
Under the default hyperparameters ($T=128,k=32$), the relative error between log-determinant estimation and the ground truth is less than $5\%$.
Moreover, our algorithm produces unbiased estimates for the derivatives with
respect to $\alpha$ and $\beta$, which enables us to confidently use stochastic
gradient methods for learning those parameters.

\xhdr{Scalability of stochastic estimation}
\begin{figure}[t]
\centering
\includegraphics[width=0.7\linewidth]{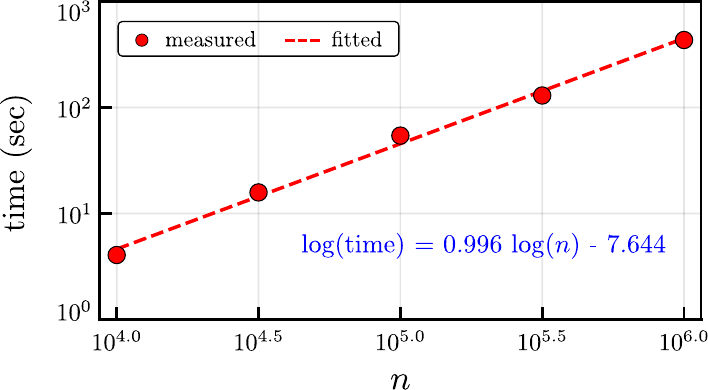}
\caption{Linear scaling of the stochastic estimation algorithm using random Watts-Strogatz graphs, where the average degree in each graph is $10$.
  Measured times are circles, and the dashed line is the linear fit (coefficients in blue).
}
\label{fig:scaling}
\end{figure}
Now, we validate the computational complexity of the proposed algorithm.
We run our algorithm on a sequence of Watts-Strogatz graphs with increasing number of vertices, with average degree fixed to be $10$.
\Cref{fig:scaling} shows that the empirical running time grows linearly with the size of the graph, as indicated by the slope of the fitted curve.


\section{Numerical Experiments \label{sec:experiments}}
Now that we have developed efficient algorithms for optimizing model parameters,
we use our model for label prediction tasks in synthetic and real-world
attributed graphs.
Our model learns both positive and negative residual correlations from
real-world data, which substantially boosts the regression accuracy and also
provides insights about the correlation among neighboring vertices.

\subsection{Data}
Our model and the baselines are tested on the following graphs (see \cref{subsec:additional_datasets} for additional datasets details).

\xhdr{Ising model}
The Ising model is a widely-used random model in statistical physics~\cite{Cipra_1987},
and we consider vertices on a $35 \times 35$ grid graph.
The spin of each vertex is either up ($+1.0$) or down ($-1.0$), which tends
to align with an external field but is also influenced by neighboring spins.
The neighboring spins are likely to be parallel if their interaction is set to be
positive, and anti-parallel otherwise.
We use Ising model samples from these two settings and denote them by Ising(+)
and Ising(-), respectively, providing synthetic datasets with clear positive and
negative correlations in labels.
We use the grid coordinates as vertex features to predict vertex spins.

\xhdr{U.S. election maps}
The election map is in \cref{fig:demo}, where
vertices are counties in the U.S.\ and edges connect bordering counties.
Each county has demographic and election statistics.%
\footnote{Graph topology and election outcomes from \url{https://github.com/tonmcg/}, other statistics from \url{www.ers.usda.gov/data-products/county-level-data-sets/}.}
We use these as both features and outcomes:
in each experiment, we select one statistic as the outcome; the remaining are vertex features.
We use 2012 and 2016 statistics.
The former is used for the transductive experiments, and both are used for the inductive experiments.

\xhdr{Transportation networks}
The transportation networks contain traffic data in the cities of Anaheim and Chicago.\footnote{Data from \url{https://github.com/bstabler/TransportationNetworks}.}
Each vertex represents a directed lane, and two lanes that meets at the same intersection are connected.
Since the lanes are directed, we create two type of edges: lanes that meets head-to-tail are connected by a type-$1$ edge, and lanes that meet head-to-head or tail-to-tail are connected by a type-$2$ edge.
For this, we use the extended model from \cref{subsec:extended_model}.
The length, capacity, and speed limit of each lanes are used as features to predict traffic flows on the lanes.
    
\xhdr{Sexual interactions}
The sexual interaction network among 1,888 individuals is from an HIV transmission study~\cite{morris2011hiv}.
We use race and sexual orientation as vertex features to predict the gender of
each person ($+1$ for male / $-1$ for female).
Most sexual interactions are heterosexual, producing negative label correlations for neighbors.
    
\xhdr{Twitch social network}
The Twitch dataset represents online friendships amongst Twitch
streamers in Portugal~\cite{rozemberczki2019multi}.
Vertex features are principal components from statistics such as the games
played and liked, location, and streaming habits.
The goal is to predict the logarithm of the number of viewers for each streamer.

\subsection{Transductive Learning \label{subsec:transductive_results}}
We first consider the transductive setting, where the training and testing vertices are from the same graph.
We find that our C-GNN framework greatly improves prediction accuracy over GNNs.

\begin{table*}[t]
  \caption{Transductive learning accuracy of our C-GNN and LP-GNN models compared to competing baselines. The best accuracy is in \bst{green}.
    Our C-GNN outperforms GNN on all datasets, often by a substantial margin.
    Even C-MLP, which does not use neighbor features, outperforms GNN in many cases, highlighting the importance of label correlation.
    LP, LP-MLP and LP-GNN assume positive label correlation among neighboring vertices and perform poorly for datasets where most edges encode negative interactions, as highlighted in \wst{orange}.
    We also report the learned $\{\alpha_i\}$ values from C-GNN.
    }
    \centering
    \resizebox{\linewidth}{!}{
   \begin{tabular}{r @{\quad} cc @{\quad} c @{\quad} ccc @{\quad} cccc}
     \toprule
    Dataset      &    $n$ &     $m$ & LP                    & MLP                    & LP-MLP                & C-MLP                 & GNN                   & LP-GNN                & C-GNN                 & $\{\alpha_{i}\}$ \\
    \midrule
    Ising(+)     & 1.2K &  2.4K & $\bst{0.76} \pm 0.02$ & $\nrm{0.68} \pm 0.03$ & $\bst{0.76} \pm 0.02$ & $\bst{0.76} \pm 0.02$ & $\nrm{0.67} \pm 0.04$ & $\bst{0.76} \pm 0.02$ & $\bst{0.76} \pm 0.02$ & $+0.89$ \\
    Ising(-)     & 1.2K &  2.4K & $\wst{0.30} \pm 0.03$ & $\nrm{0.47} \pm 0.02$ & $\wst{0.30} \pm 0.03$ & $\bst{0.77} \pm 0.03$ & $\nrm{0.47} \pm 0.03$ & $\wst{0.30} \pm 0.03$ & $\bst{0.77} \pm 0.03$ & $-0.93$ \\
    \midrule
    income       & 3.2K & 12.7K & $\nrm{0.54} \pm 0.04$ & $\nrm{0.64} \pm 0.03$ & $\nrm{0.73} \pm 0.03$ & $\nrm{0.74} \pm 0.03$ & $\nrm{0.75} \pm 0.03$ & $\bst{0.81} \pm 0.03$ & $\bst{0.81} \pm 0.02$ & $+0.92$ \\
    education    & 3.2K & 12.7K & $\nrm{0.36} \pm 0.05$ & $\nrm{0.67} \pm 0.03$ & $\nrm{0.71} \pm 0.02$ & $\bst{0.72} \pm 0.02$ & $\nrm{0.70} \pm 0.02$ & $\bst{0.72} \pm 0.03$ & $\bst{0.72} \pm 0.03$ & $+0.78$ \\
    unemployment & 3.2K & 12.7K & $\nrm{0.70} \pm 0.03$ & $\nrm{0.43} \pm 0.05$ & $\nrm{0.69} \pm 0.04$ & $\nrm{0.77} \pm 0.03$ & $\nrm{0.55} \pm 0.04$ & $\nrm{0.75} \pm 0.05$ & $\bst{0.78} \pm 0.03$ & $+0.99$ \\
    election     & 3.2K & 12.7K & $\nrm{0.58} \pm 0.02$ & $\nrm{0.37} \pm 0.02$ & $\nrm{0.61} \pm 0.03$ & $\nrm{0.63} \pm 0.03$ & $\nrm{0.51} \pm 0.04$ & $\bst{0.69} \pm 0.03$ & $\bst{0.69} \pm 0.03$ & $+0.95$ \\
    \midrule
    Anaheim      &  914 &  3.8K & $\nrm{0.49} \pm 0.08$ & $\nrm{0.75} \pm 0.02$ & $\nrm{0.81} \pm 0.04$ & $\bst{0.82} \pm 0.03$ & $\nrm{0.76} \pm 0.03$ & $\nrm{0.81} \pm 0.04$ & $\bst{0.82} \pm 0.03$ & $+0.95, +0.17$ \\
    Chicago      & 2.2K & 15.1K & $\nrm{0.59} \pm 0.05$ & $\nrm{0.60} \pm 0.05$ & $\nrm{0.65} \pm 0.06$ & $\nrm{0.65} \pm 0.05$ & $\nrm{0.68} \pm 0.04$ & $\bst{0.72} \pm 0.04$ & $\nrm{0.71} \pm 0.04$ & $+0.85, +0.68$ \\
    \midrule
    sexual       & 1.9K &  2.1K & $\wst{0.37} \pm 0.06$ & $\nrm{0.68} \pm 0.02$ & $\wst{0.64} \pm 0.03$ & $\nrm{0.83} \pm 0.03$ & $\nrm{0.88} \pm 0.02$ & $\wst{0.86} \pm 0.02$ & $\bst{0.93} \pm 0.01$ & $-0.98$ \\
    \midrule
    Twitch-PT    & 1.9K & 31.3K & $\nrm{0.00} \pm 0.04$ & $\nrm{0.61} \pm 0.03$ & $\nrm{0.60} \pm 0.04$ & $\nrm{0.66} \pm 0.03$ & $\nrm{0.69} \pm 0.03$ & $\nrm{0.69} \pm 0.03$ & $\bst{0.74} \pm 0.03$ & $+0.99$ \\
    \bottomrule
   \end{tabular}
   }
    \label{tab:trans_accuracy}
\end{table*}
\begin{figure}[t]
\centering
\includegraphics[width=0.80\linewidth]{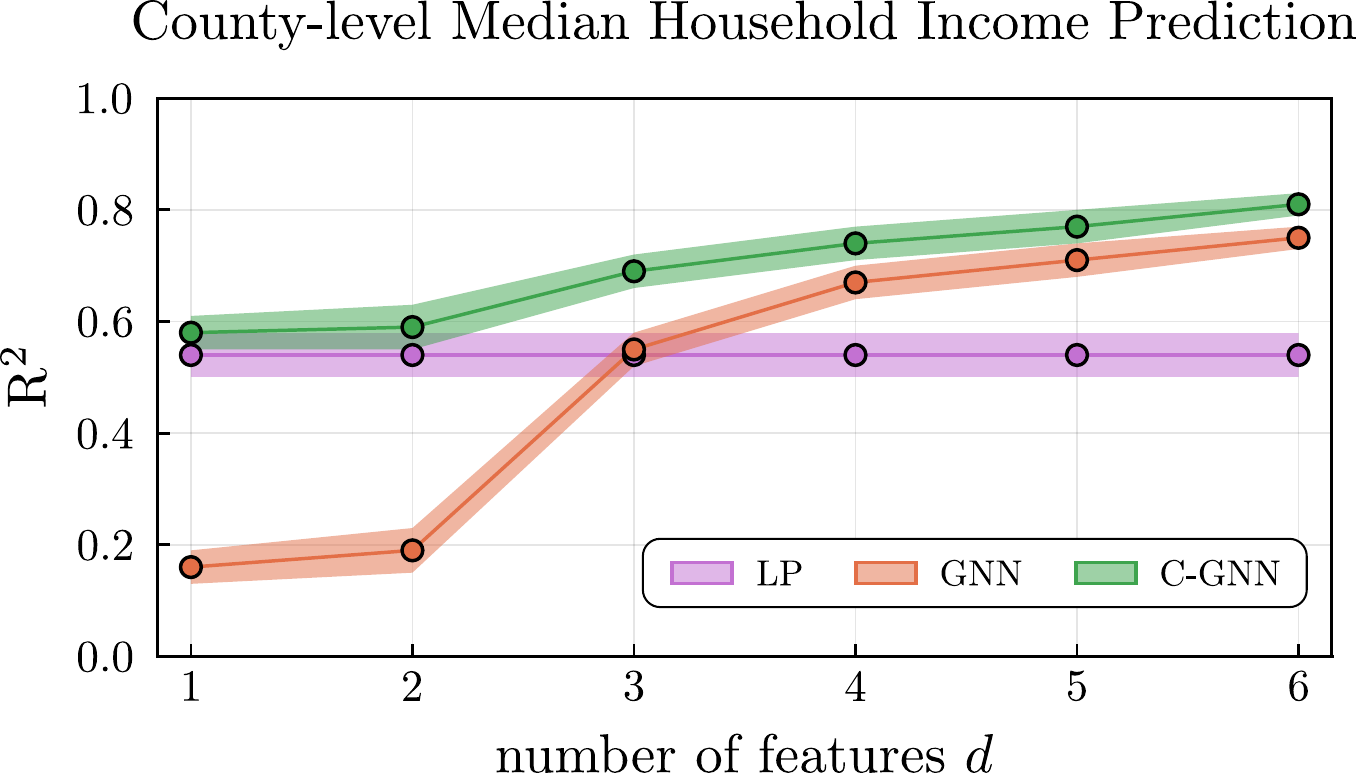}
\caption{Transductive learning accuracy for county-level median household incomes, as a function of the number of included features.
  Label propagation (LP) can work well with few features, while the GNN can work well with many features;
  however, C-GNN outperforms both in all cases.
}
\label{fig:crossover}
\end{figure}

\xhdr{Methods and baselines}
We use a 2-layer GraphSAGE GNN with ReLU activations and mean-aggregation~\cite{Hamilton_2017}  as the base predictor in our framework.
(Other GNN architecture provide similar results; see \cref{subsec:other_gnns}.)
We compare C-GNN against label propagation (LP)~\cite{Zhu_2003}, a multi-layer
perceptron (MLP; architecture details in \cref{subsec:additional_setup}), the
base GNN, and the LP-GNN algorithm from \cref{subsec:lp_gnn}.
LP assumes and takes advantage of positive label correlation among neighboring vertices, but it does not use vertex features.
On the other hand, the MLP ignores the label correlations, and only uses the features of a given vertex to predict its label.
We also tested our correlation framework with the MLP as the base regressor instead of
the GNN (C-MLP and LP-MLP).

\xhdr{Setup and performance metric}
For each graph, we normalize each vertex feature to have zero mean and
unit standard deviation, and randomly split the vertices into $60\%$ for training,
$20\%$ for validation, and $20\%$ for testing.
The GNN parameters are trained using the ADAM optimizer with default learning rate,
while the model parameters $\alpha, \beta$ are optimized with gradient descent
along with the GNN parameters.
For the Ising model and sexual interaction datasets, the vertex
labels are binary, so we threshold the regression output at 0 and use
binary classification accuracy as the performance metric.
For the rest of datasets, we use coefficients of determination $R^{2}$ to measure accuracy.
Each combination of method and dataset is repeated 10 times with different
random seeds, and the mean and standard deviation of the accuracies are
recorded.

\xhdr{Main results}
\Cref{tab:trans_accuracy} summarizes the results.
C-GNN substantially improves the prediction accuracy over GNN for all datasets:
the C-GNN mean classification accuracy is $0.82$ over the Ising spin and sexual interaction datasets, and the mean $R^{2}$ is $0.75$ over the remaining datasets,
while the GNN mean classification and $R^{2}$ accuracies were $0.67$ and $0.66$, respectively.
Moreover, our LP-GNN also performs very well on most of the datasets,
with performance on par with C-GNN in five datasets and performing at least as well as the standard GNN in 8 out of 10 datasets.
The two datasets on which it performs poorly are Ising(-) and the sexual interaction network, where the labels of connected vertices are likely to be negatively correlated;
this is expected since the LP-GNN model assumes positive correlations between neighbors.
Interestingly, our framework also significantly improves the performance of the MLP.
In fact, C-MLP is often much better than a standard GNN.
This is evidence that oftentimes more performance can be gained from modeling
label correlation as opposed to sophisticated feature aggregation.

The learned parameters also reveal interaction types.
The learned $\{\alpha_{i}\}$ are all positive except for the Ising(-) and sexual interaction
datasets, where the vertex labels are negatively correlated.
Moreover, for the traffic graph, the learned $\alpha_{1} > \alpha_{2}$ indicates
that traffic on two lanes with head-to-tail connection are more strongly correlated,
since a vehicle can directly move from one lane to another.

\xhdr{Understanding performance better}
We perform a more in-depth analysis for predicting county-level median household income.
This dataset has six features (migration rate, birth rate, death rate, education level, unemployment rate, and election outcome),
and we use the first $d$ for income prediction, comparing against LP and GNN (\cref{fig:crossover}).
The GNN performs poorly for small $d$, but gradually surpasses LP as more features are available.
Our C-GNN method outperforms both regardless of $d$, although the performance
gap between C-GNN and GNN narrows as $d$ increases.
These results highlight that, if features are only mildly predictive, accounting for
label correlation can have an enormous benefit.

\subsection{Inductive Learning \label{subsec:ind_results}}
\begin{figure}[t]
\centering
\includegraphics[width=1.00\linewidth]{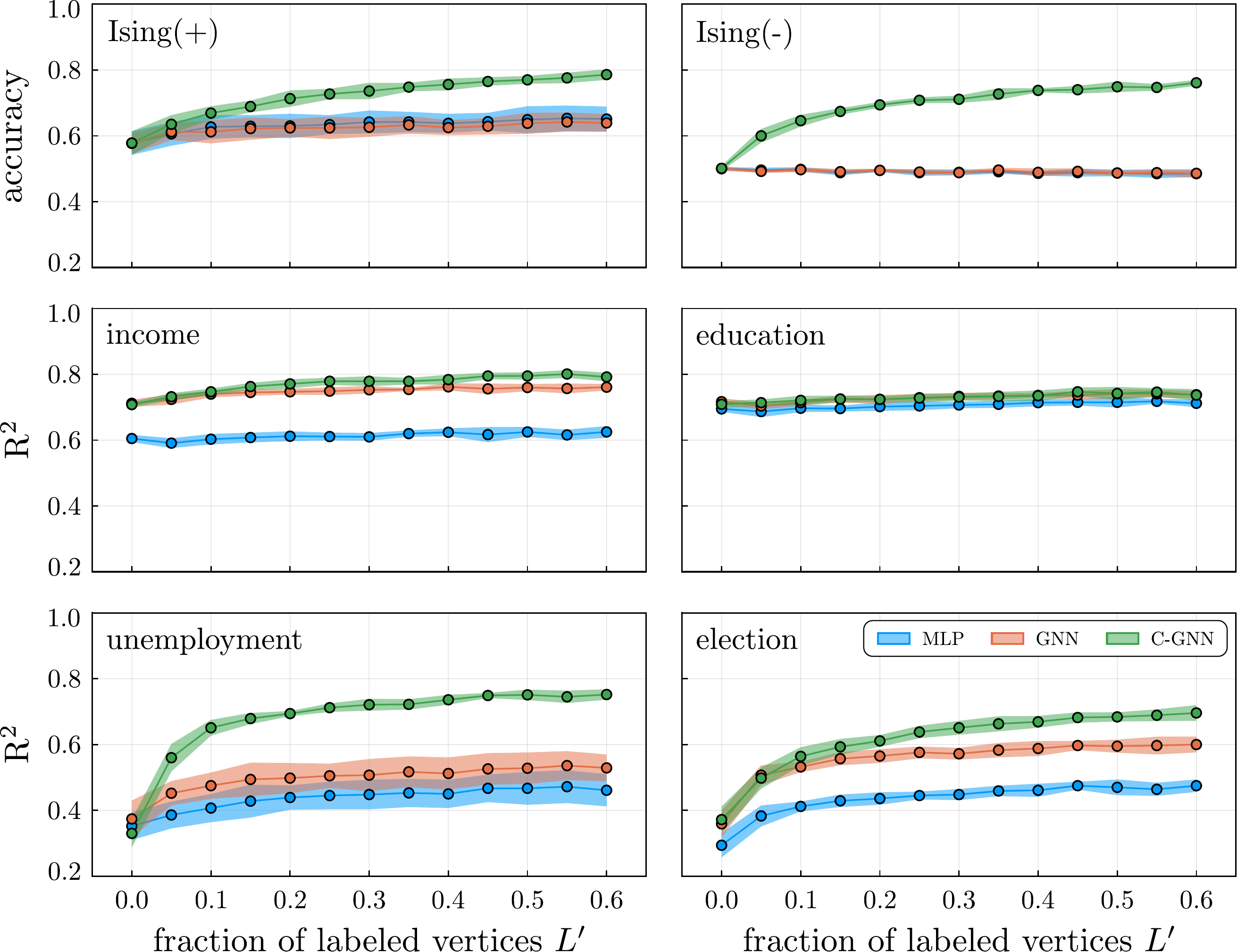}
\caption{Inductive learning accuracy with a fraction of labeled vertices in the new graph.
C-GNN is able to utilize the extra labels more effectively than the baselines, and it does not need neural network fine-tuning.
}
\label{fig:ind}
\end{figure}
We now consider the inductive setting, where a model is trained on vertex labels
from one graph $G$ and tested on an unseen graph $G'$.
This setting is useful when both graphs have similar properties,
but vertex labels in $G'$ are more expensive to obtain.
In particular, we consider the scenario where a small fraction of vertex labels
in $G'$ are available and demonstrate how our framework allows using those extra
labels to improve regression accuracy.
We denote the labeled and unlabeled vertices in $G'$ as $L'$ and $U'$.

\xhdr{Datasets and methods}
We use the Ising model and election map datasets for inductive learning experiments.
In the former, the spin configurations on $G$ and $G'$ are simulated under the same Ising model setting.
In the latter, we train with the 2012 data and test on the 2016 election map, predicting several attributes.
We compare C-GNN to GNN and MLP.
The C-GNN is trained using $60\%$ vertex labels from $G$, and tested directly on $U'$ by conditioning on the vertex labels of $L'$.
The GNN and MLP are first trained on $G$;
for a fair comparison, we then use the learned parameters as the
initial guess for $G'$, and optimize the model further with the labels on $L'$.

\xhdr{Results}
We test the performance of our framework and the baselines for different sizes of $L'$ (\cref{fig:ind}).
C-GNN and GNN gives the same prediction accuracy if no vertex label on $G'$ is available,
but, as the number of labeled points increases, C-GNN outperforms the baselines by large margins on multiple datasets.
This indicates that the learned residual correlation generalizes well to unseen graphs.
Household income and education level predictions do not benefit much from our
framework, partially because those statistics are relatively stable over time,
so the models trained on $2012$ data are already a good proxy for $2016$.
Remarkably, C-GNN works well without fine-tuning the neural network parameters
on the new labels of $L'$, indicating that the feature-label mapping oftentimes
shifts from $G$ to $G'$ collectively amongst neighboring vertices.


\section{Related Work \label{sec:related_works}}
By now, semi-supervised learning on graphs has been extensively studied~\cite{Zhu_2003,Zhou_2004,ibrahim2019nonlinear,Jia_2019}.
Label propagation or diffusion ``distributes'' observed vertex labels throughout
the graph~\cite{Zhu_2003,Zhou_2004}, but were not designed to incorporate
additional vertex features.
Laplacian Regularization~\cite{Ando_2006} and Manifold
regularization~\cite{Belkin_2006} propose to augment feature-based supervised
learning methods with an unsupervised loss function that minimize differences
between connected vertices.
These methods assume neighboring vertices should have similar labels, which is
true in many applications.

There are direct models of correlation structure for graph-based semi-supervised learning~\cite{Xu_2010};
such approaches are more computationally expensive and not amenable to joint learning with GNN parameters.
The marginalized Gaussian conditional random field
(m-GCRF)~\cite{Stojanovic_2015} is closer to our approach, using a CRF to model
the label distribution given the vertex features, which reduces to Gaussian
distributions with sparse precision matrices under the right choice of potential
function.
In contrast, we model the correlation of regression residuals instead of the outcomes themselves,
and our precision matrix parameterization enables linear-time learning.

The inability of existing GNN approaches to capture label correlations has been
discussed in the classification setting.
Recent proposals include graph Markov neural networks~\cite{Qu_2019} and conditional
graph neural fields~\cite{Gao_2019}, which use a CRF to model the joint
distribution of vertex classes; as well as positional
GNNs~\cite{you2019position}, which use a heuristic of letting GNN aggregation
parameters depend on distances to anchor nodes.
With the CRF approaches, the joint likelihood does not have a closed form
expression, and such models are trained by maximizing the pseudo-likelihood with
the expectation-maximization algorithm.
The regression setting here is more mathematically convenient: an unbiased exact
joint likelihood estimate can be quickly computed, and the outcome has an
interpretable decomposition into base prediction and residual.


\section{Discussion \label{sec:conclusion}}
Our semi-supervised regression framework combines the advantages of GNNs and
label propagation to get value from both vertex feature information and
outcome correlations.
Our experiments show that accounting for outcome correlations can give enormous
performance gains, especially in cases where the base prediction by a GNN is
only mildly accurate.
In other words, label correlations can provide information complementary
(rather than redundant) to vertex features in some datasets.
Understanding this more formally is an interesting avenue for future research.

Our C-GNN uses only a few parameters to model the label correlation structure,
and learns the direction and strength of correlations with highly efficient
algorithms.
The model also enables us to measure uncertainty in predictions, although
we did not focus on this.
The C-GNN can model more types of data and requires some careful numerical algorithms
to scale well.
Our simplified LP-GNN approach offers a simple, light-weight add-on to any GNN
implementation that can often substantially boost performance.

\section*{Acknowledgements}
This research was supported by NSF award DMS-1830274; ARO award W911NF-19-1-0057; ARO MURI; and JPMorgan Chase \& Co.

\bibliographystyle{ACM-Reference-Format}
\bibliography{main}
\appendix
\clearpage

\section{Appendix \label{sec:supplementary}}
Here we provide some implementation details of our methods to help readers reproduce and further understand the algorithms and experiments in this paper.
All of the algorithms used in this paper are implemented in Julia 1.2.
The source code, data, and experiments are available at: \\
\centerline{\url{https://github.com/000Justin000/gnn-residual-correlation}.}
\subsection{Label Propagation Algorithm \label{subsec:label_propagation}}
Given targets on the training vertices $\mathbf{z}_{L}$, LP computes the targets on the testing vertices $\mathbf{z}_{U}$ with the following constrained minimization:
\begin{align}
    \mathbf{z}^{\rm LP} = \arg\min_{\hat{\mathbf{z}}} \hat{\mathbf{z}}^{\intercal} \mathcal{L} \hat{\mathbf{z}} \qquad \textrm{s.t.} \qquad \hat{\mathbf{z}}_{L} = \mathbf{z}_{L}
    \label{eq:lp_objective}
\end{align}
where $\mathcal{L} = \mathbf{I} - \mathbf{S}$ is the normalized Laplacian matrix.
This is the method by Zhu et al.~\cite{Zhu_2003} but with the
normalized Laplacian instead of the combinatorial Laplacian, 
which is nearly the same as the approach by Zhou et al.~\cite{Zhou_2004},
except targets on $L$ are fixed.
The solution on the unlabeled vertices is
\begin{align}
    \mathbf{z}_{U}^{\rm LP} = - \mathcal{L}_{UU}^{-1} \mathcal{L}_{UL} \mathbf{z}_{L},
    \label{eq:lp_solution}
\end{align}
which we can compute with CG.
If $L$ and $U$ are disconnected, $\mathcal{L}_{UU}$ is singular. Then starting with an all-zero initial guess, CG converges to the minimal norm solution that satisfies \cref{eq:lp_objective}~\cite{Hayami_2018}.
The entire algorithm is summarized in \cref{alg:lp}.
\begin{algorithm}[h]
\SetKwInOut{Input}{Input}
\SetKwInOut{Output}{Output}
\Input{\ normalized adjacency matrix $\mathbf{S}$; training targets $\mathbf{z}_{L}$ (label or residual); }
\Output{\ predicted targets $\mathbf{z}_{U}^{\rm LP}$ for unknown vertices}
\BlankLine
$\mathcal{L} \gets \mathbf{I} - \mathbf{S}$ \Comment{precision matrix} \\
$\mathbf{z}_{U}^{0} \gets \mathbf{0}$ \Comment{initial guess} \\
$\mathbf{z}_{U}^{\rm LP} \gets \texttt{ConjugateGradient}(\mathcal{L}_{UU}, -\mathcal{L}_{UL} \mathbf{z}_{L}, \mathbf{z}_{U}^{0})$ \\
\caption{Label Propagation.}
\label{alg:lp}
\end{algorithm}

\subsection{Stochastic logdet Estimation with \texttt{Flux.jl} \label{subsec:flux_implementation}}
The base GNN regressors are implemented in Julia with \texttt{Flux.jl}~\cite{Innes_2018}.
For better compatibility with the underlying GNN, we implement the stochastic estimation algorithm using the ``customized gradient'' interface provided by \texttt{Flux.jl}.
For example, \cref{lst:logdet} shows the code snippet that defines the log-determinant computation:
when \texttt{logdet$\Gamma$} is invoked, its output is tracked and its derivative can be computed automatically with back-propagation.
This scheme greatly simplifies the downstream implementations for data mining
experiments, and it works as if we were computing the exact gradient --- only
orders of magnitude faster with a minor loss of accuracy, as evidenced
by our experiments in \cref{sec:approximation_validation}.
\begin{listing}[h]
\begin{minted}{julia}
using Flux.Tracker: track, @grad

# When this function is invoked, Flux automatically 
# tracks the output for auto-differentiation
logdetΓ(α, β; S, t, k) = track(logdetΓ, α, β; S=S, t=t, k=k);

# This tells Flux how to track logdetΓ
@grad function logdetΓ(α, β; S, t, k)
    """ 
    Input:
      α: (vector of) model parameters
      β: model parameter
      S: (vector of) normalized adjacency matrices
      t: # of trial vectors
      k: # of Lanczos tridiagonal iterations

    Output:
      1): logdet(Γ)
      2): map from sensitivity of logdet(Γ) 
                to sensitivity of α, β
    """
    # sample Gaussian random vector
    n = size(S,1);
    Z = randn(n,t);

    # eqns (5) in this paper
    Γ = getΓ(α, β; S=S);
    ∂Γ∂α = get∂Γ∂α(α, β; S=S);
    ∂Γ∂β = get∂Γ∂β(α, β; S=S);

    # adopted from Gardner 2018 GPytorch paper
    X, TT = mBCG(Y->Γ[P,P]*Y, Z; k=k);

    # eqn (20) in this paper
    vv = 0;
    for T in TT
        eigvals, eigvecs = eigen(T);
        vv += sum(eigvecs[1,:].^2 .* log.(eigvals));
    end 
    Ω = vv*n/t;

    # eqn (18) in this paper
    trΓiM(M) = sum(X.*(M[P,P]*Z))/t;
    ∂Ω∂α = map(trΓiM, ∂Γ∂α);
    ∂Ω∂β = trΓiM(∂Γ∂β);

    return Ω, Δ -> (Δ*∂Ω∂α, Δ*∂Ω∂β);
end
\end{minted}
\caption{Code snippet for estimating the log-determinant and its derivatives.}
\label{lst:logdet}
\end{listing}
\subsection{Additional Details on Experimental Setup\label{subsec:additional_setup}}
\xhdr{Neural network architecture}
Our regression pipeline first encodes each vertex into an 8-dimension representation using an MLP or GNN and then uses a linear output layer to predict its label.
For the MLP, we use a 2-hidden-layer feedforward network with 16 hidden units and ReLU activation function.
Each GNN we consider also consists of 2 layers, each with 16 hidden units and ReLU activation function.

\xhdr{Optimization}
For all but the Twitch-PT datasets, the framework parameters $\alpha, \beta$ are optimized using gradient descent with learning rate $10^{-1}$.
The Twitch-PT dataset uses a limited-memory BFGS optimizer.
For the MLP and GNN experiments summarized in \cref{tab:trans_accuracy,fig:crossover}, the neural network parameters are optimized for 75 epochs using the Adam optimizer
with $\beta_{1} = 0.9$, $\beta_{2} = 0.999$, and learning rate $10^{-3}$.
For the inductive experiments summarized in \cref{fig:ind}, the neural network parameters are further fine-tuned for 25 epochs with the Adam optimizer at a smaller learning rate $5 \times 10^{-4}$.
All of our experiments are performed on a single workstation with an 8-core i7-7700 CPU @ 3.60GHz processor and 32 GB memory. 

\begin{figure*}[ht]
\centering
\includegraphics[width=0.85\linewidth]{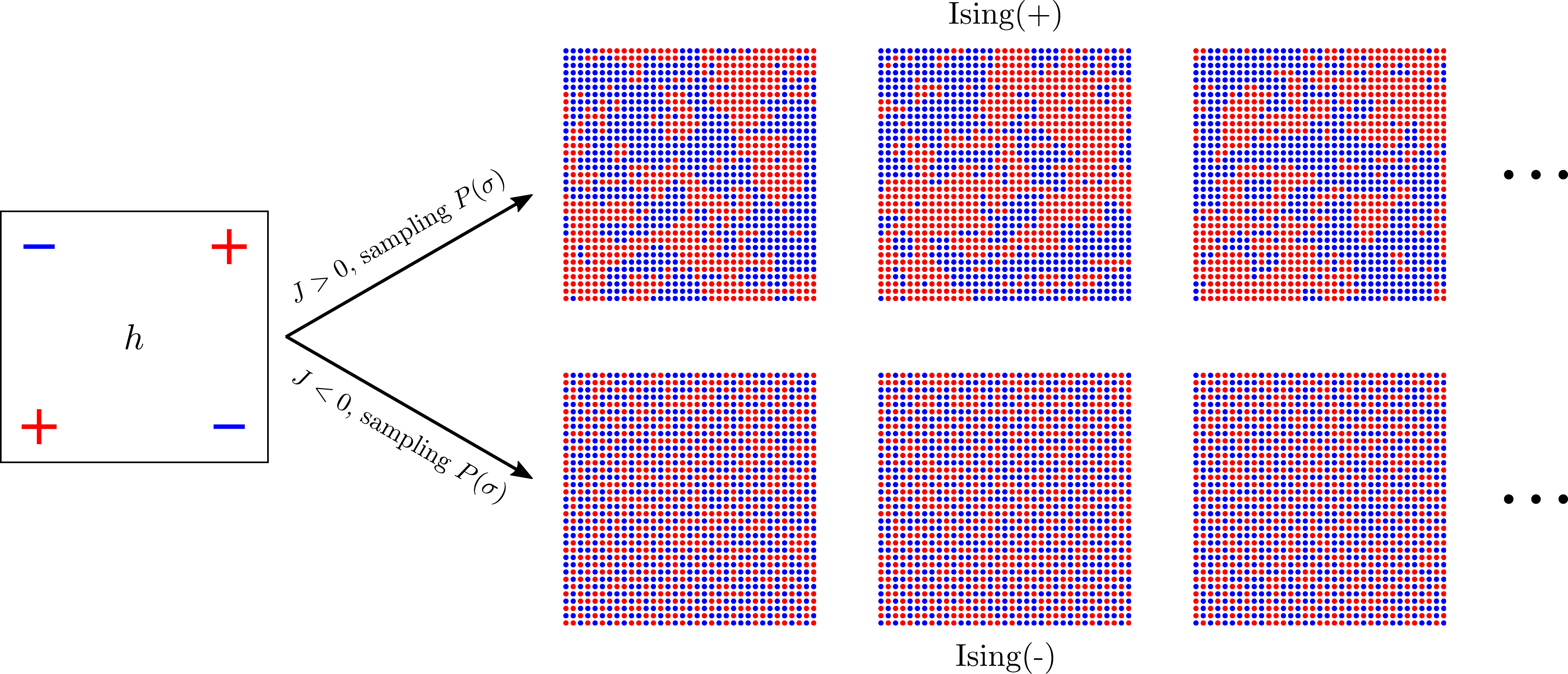}
\caption{Example Ising spin configurations sampled from the Boltzmann distribution. Vertices with $+1$ spins are colored in red, and vertices with $-1$ spin are colored in blue.
}
\label{fig:ising}
\end{figure*}

\begin{table*}[ht]
    \centering
    \caption{Transductive learning accuracy of our framework, using different GNN architecture as the base predictor.}
    \resizebox{\linewidth}{!}{
    \begin{tabular}{rccccccccc}
    \toprule
                 & \multicolumn{3}{c}{GCN~\cite{Kipf_2016}}                                              & \multicolumn{3}{c}{GraphSAGE-max~\cite{Hamilton_2017}}                                    & \multicolumn{3}{c}{GraphSAGE-pooling~\cite{Hamilton_2017}} \\ \cmidrule(lr){2-4}\cmidrule(lr){5-7}\cmidrule(lr){8-10}
    Dataset      & GNN                    & LP-GNN                & C-GNN                & GNN                   & LP-GNN                & C-GNN                 & GNN                   & LP-GNN                & C-GNN                 \\
    \midrule                                                                                                                                                        
    Ising(+)     & $\nrm{0.61} \pm 0.04$ & $\nrm{0.72} \pm 0.03$ & $\nrm{0.72} \pm 0.03$ & $\nrm{0.67} \pm 0.04$ & $\nrm{0.76} \pm 0.02$ & $\nrm{0.76} \pm 0.02$ & $\nrm{0.67} \pm 0.04$ & $\nrm{0.76} \pm 0.02$ & $\nrm{0.76} \pm 0.02$ \\
    Ising(-)     & $\nrm{0.47} \pm 0.02$ & $\nrm{0.34} \pm 0.02$ & $\nrm{0.70} \pm 0.03$ & $\nrm{0.47} \pm 0.02$ & $\nrm{0.30} \pm 0.03$ & $\nrm{0.77} \pm 0.03$ & $\nrm{0.48} \pm 0.02$ & $\nrm{0.30} \pm 0.03$ & $\nrm{0.77} \pm 0.02$ \\
    \midrule                                                                                                                                                        
    income       & $\nrm{0.60} \pm 0.04$ & $\nrm{0.61} \pm 0.05$ & $\nrm{0.62} \pm 0.04$ & $\nrm{0.73} \pm 0.03$ & $\nrm{0.79} \pm 0.03$ & $\nrm{0.78} \pm 0.04$ & $\nrm{0.74} \pm 0.03$ & $\nrm{0.78} \pm 0.03$ & $\nrm{0.77} \pm 0.02$ \\
    education    & $\nrm{0.45} \pm 0.04$ & $\nrm{0.44} \pm 0.04$ & $\nrm{0.47} \pm 0.04$ & $\nrm{0.67} \pm 0.02$ & $\nrm{0.70} \pm 0.02$ & $\nrm{0.70} \pm 0.03$ & $\nrm{0.68} \pm 0.02$ & $\nrm{0.70} \pm 0.03$ & $\nrm{0.70} \pm 0.03$ \\
    unemployment & $\nrm{0.49} \pm 0.03$ & $\nrm{0.72} \pm 0.03$ & $\nrm{0.72} \pm 0.03$ & $\nrm{0.57} \pm 0.05$ & $\nrm{0.74} \pm 0.04$ & $\nrm{0.75} \pm 0.05$ & $\nrm{0.60} \pm 0.05$ & $\nrm{0.74} \pm 0.04$ & $\nrm{0.75} \pm 0.04$ \\
    election     & $\nrm{0.45} \pm 0.03$ & $\nrm{0.61} \pm 0.02$ & $\nrm{0.60} \pm 0.02$ & $\nrm{0.43} \pm 0.04$ & $\nrm{0.64} \pm 0.03$ & $\nrm{0.65} \pm 0.03$ & $\nrm{0.49} \pm 0.06$ & $\nrm{0.66} \pm 0.02$ & $\nrm{0.65} \pm 0.02$ \\
    \midrule                                                                                                                                                        
    Anaheim      & $\nrm{0.69} \pm 0.05$ & $\nrm{0.75} \pm 0.05$ & $\nrm{0.75} \pm 0.05$ & $\nrm{0.73} \pm 0.04$ & $\nrm{0.79} \pm 0.05$ & $\nrm{0.80} \pm 0.04$ & $\nrm{0.74} \pm 0.04$ & $\nrm{0.80} \pm 0.04$ & $\nrm{0.80} \pm 0.05$ \\
    Chicago      & $\nrm{0.58} \pm 0.05$ & $\nrm{0.63} \pm 0.05$ & $\nrm{0.63} \pm 0.05$ & $\nrm{0.64} \pm 0.05$ & $\nrm{0.68} \pm 0.05$ & $\nrm{0.68} \pm 0.05$ & $\nrm{0.66} \pm 0.05$ & $\nrm{0.69} \pm 0.04$ & $\nrm{0.68} \pm 0.04$ \\
    \midrule                                                                                                                                                        
    sexual       & $\nrm{0.77} \pm 0.04$ & $\nrm{0.72} \pm 0.04$ & $\nrm{0.92} \pm 0.02$ & $\nrm{0.86} \pm 0.02$ & $\nrm{0.86} \pm 0.02$ & $\nrm{0.92} \pm 0.02$ & $\nrm{0.85} \pm 0.05$ & $\nrm{0.85} \pm 0.04$ & $\nrm{0.92} \pm 0.02$ \\
    \midrule                                                                                                                                                        
    Twitch-PT    & $\nrm{0.54} \pm 0.02$ & $\nrm{0.65} \pm 0.01$ & $\nrm{0.64} \pm 0.02$ & $\nrm{0.69} \pm 0.04$ & $\nrm{0.69} \pm 0.04$ & $\nrm{0.70} \pm 0.03$ & $\nrm{0.72} \pm 0.03$ & $\nrm{0.72} \pm 0.03$ & $\nrm{0.71} \pm 0.03$ \\
    \bottomrule
    \end{tabular}
    }
    \label{tab:trans_accuracy_other_gnn}
\end{table*}

\subsection{Additional Details on Datasets \label{subsec:additional_datasets}}
\xhdr{Ising model simulations}
The Ising model samples random labels on a two-dimensional grid graph.
For each vertex $i$, there is a discrete variable $\sigma_{i} \in \{-1, +1\}$ representing its spin state.
A spin configuration $\sigma$ assigns a spin state to every vertex in the graph.
The Ising model considers two type of interactions: (i) interaction between the spin of each vertex with external field and 
(ii) the interaction between neighboring spins.
Those interactions constitute the ``energy'' for each spin configuration:
\begin{align}
    H(\sigma) = -\sum_{(i,j) \in E} J_{ij} \sigma_{i} \sigma_{j} - \sum_{i \in V} h_{i} \sigma_{i},
\end{align}
where $J_{ij}$ controls the interaction between neighboring vertices, and $h_{i}$ denotes the external field on vertex $i$.
Finally, the configuration probability is given by the Boltzmann distribution,
\begin{align}
    P(\sigma) = \frac{e^{-H(\sigma)}}{\sum_{\sigma'} e^{-H(\sigma')}}.
\end{align}
Our Ising spin simulation randomly draws from this Boltzmann distribution.
For the Ising(+) dataset, we set $J_{ij} = J = 0.1$ and $h_{i} = 0.35 \cdot (x_{i})_{1} \cdot (x_{i})_{2}$, where $\mathbf{x}_{i}$ is the coordinate of vertex $i$ normalized between $-1.0$ and $+1.0$.
In other words, the system favors parallel spins between neighboring vertices, and the external field exhibits an ``XNOR'' spatial pattern.
For the Ising(-) dataset, a similar setting is used, except that $J_{ij} = J = -0.1$.
Some sampled Ising spin configurations are shown in \cref{fig:ising}.

\xhdr{Sexual interaction dataset}
The dataset used to construct the sexual interaction network was collected by Colorado Springs project 90, which details the relationships of 7,674 individuals.
We take the largest connected component in the derived sexual relation network, which consists of 1,888 vertices and 2,096 edges.
Of the 2,096 relationships, 2,007 are heterosexual and 89 are homosexual.

\subsection{Other GNN Base Predictors \label{subsec:other_gnns}}
We tested a variety of GNN architectures as base regressors in our framework,
and \cref{tab:trans_accuracy_other_gnn} summarizes the results.
Here, we see the exact same trend as describe in
\cref{subsec:transductive_results}: C-GNN substantially out-performs the base GNN on almost all
datasets, and LP-GNN outperforms GNN on datasets where vertex labels are
positively correlated.
These experiments support our claim that the performance gains we observe from
exploiting label correlation is robust to change of the underlying GNN
architecture.

\end{document}